# Probability Logic

Niki Pfeifer

University of Regensburg

Department of Philosophy

Universitätsstraße 31

D-93053 Regensburg

niki.pfeifer@ur.de

**Short Abstract**

This chapter presents probability logic as a rationality framework for human reasoning under uncertainty. Selected formal-normative aspects of probability logic are discussed in the light of experimental evidence. Specifically, probability logic is characterized as a generalization of bivalent truth-functional propositional logic (short "logic"), as being connexive, and as being nonmonotonic. The chapter discusses selected argument forms and associated uncertainty propagation rules. Throughout the chapter, the descriptive validity of probability logic is compared to logic, which was used as the gold standard of reference for assessing the rationality of human reasoning in the 20$^{th}$ century.

## Probability logic is a generalization of logic

Probability logic as a rationality framework combines probabilistic reasoning with logical rule-based reasoning and studies formal properties of uncertain argument forms. Among various approaches to probability logic (for overviews see, e.g., Hailperin, 1996; Adams, 1975, 1998; Coletti and Scozzafava, 2002; Haenni, Romeijn, Wheeler, and Williamson, 2011; Demey, Kooi, and Sack, 2017), this chapter reviews selected formal-normative aspects of probability logic in the light of experimental evidence. The focus is on probability logic as a generalization of the classical propositional calculus (short: logic; for probabilistic generalizations of quantified statements see, e.g., Hailperin, 2011; Pfeifer & Sanfilippo, 2017, 2019). The generalization is obtained by (i) the use of probability functions and (ii) by the introduction of the conditional event as a logical object, which is not expressible within logic. This generalization is currently most frequently investigated from a psychological point of view (see, e.g., Pfeifer and Kleiter, 2009; Oaksford and N. Chater, 2010; Elqayam, Bonnefon, and Over, 2016) and it is thus most suitable for discussing empirical and normative aspects of probability logic as a rationality framework. The empirical focus is on investigating general patterns of human reasoning under uncertainty and not on data modeling.

Logic is bivalent as it deals with *true* and *false* as the two truth values, which are assigned to propositional variables (denoted by upper case letters *in italics*; for logic see also chapter 3.1. by Steinberger, in this volume). Truth-tables can be used to define logical connectives, like *conjunction* ("*A* and *B*", denoted by "*A&B*"), *disjunction* ("*A* or *B*", denoted by "*AvB*"), *negation* ("not-*A*" denoted by "~*A*"), or the *material conditional*, which is the disjunction ~*AvB*. Probability logic generalizes logic by using the real-valued unit interval from zero to one instead of just two truth values: truth value functions are generalized by probability functions. While logic is truth-functional, probability logic is only partially truth functional as usually a probability-interval is obtained in the conclusion even from point-valued premise probabilities. Probability functions can be used to formalize a real or an ideal agent's degree of belief in propositions formed by logical connectives, e.g., in *A&B, AvB,* or in ~*A*. Moreover, conditional probability functions can measure the degree of belief in a *conditional event*, i.e., *p*(*B*|*A*). The conditional event *B*|*A* is a three-valued logical entity, which is *true* if *A&B* is true, *false* if *A&*~*B* is true, and *void* (or undetermined) if ~*A* is true.[1] Since the conditional event cannot be expressed by a two-valued proposition, it is by definition not propositional and constitutes a further generalization of logic. Using the betting interpretation of probability, "true" means that the bet is won, "false" means that the bet is lost, and "void" means that the bet is called off (i.e., you get your money back).

---

1   Note that the conditional event must not be nested: neither *A* nor *B* in *B*|*A* may contain occurrences of "|", because of Lewis' (1976) triviality results. For nested conditionals and logical operations among conditionals complexer structures are needed to avoid triviality,

Note that $p(A)=0$ does not imply that $A$ is logically impossible (i.e., a logical contradiction $\bot$). However, $p(\bot)$ is necessarily equal to zero. As it does not make sense to add $\bot$ to your stock of belief (or to bet on a conditional where $\bot$ is its antecedent), it is obvious why "$A|\bot$" is undefined in the coherence approach.[2] The semantics of the conditional event matches the participant's responses in the truth table tasks (see, e.g., Wason & Johnson-Laird, 1972; see also chapter 4.6. by Oberauer and Pessach in this volume): most people judge that (i) $A\&B$ confirms the conditional *if A then B,* (ii) $A\&\sim B$ disconfirms it, and that (iii) $\sim A$ is irrelevant for *if A then B*. Under the material conditional interpretation one would expect that rational people judge that $\sim A$ confirms the conditional. As this expectation was violated, the response pattern (i)-(iii) was pejoratively called "defective truth table". Within the rationality framework of probability logic, however, this response is perfectly rational, as it matches the semantics of the conditional event (see, e.g., Over and Baratgin, 2017; Pfeifer and Tulkki, 2017b; Kleiter et al., 2018; see also chapter 6.2. by Over and Cruz in this volume).

**Conditional probability, zero-antecedent probabilities, and paradoxes**

Traditionally, conditional probability is defined by

(1) $p(B|A) =_{\text{def.}} p(A\&B)/p(A)$, if $p(A)>0$.

Condition $p(A)>0$ serves here to avoid fractions over zero. But what if $p(A)=0$? Then, the conditional probability is undefined or default assumptions about $p(B|A)$ are made. Some

---

which go beyond the scope of this chapter (see, e.g., the theory of "conditional random quantities"; Gilio and Sanfilippo, 2014; Sanfilippo, Pfeifer, & Gilio, 2017; Sanfilippo, Pfeifer, Over, & Gilio, 2018).
2 Popper functions, however, allow for conditioning on contradictions (see Coletti et al., 2001, for a discussion).

approaches suggest for example, by default, to equate $p(B|A)$ with 1 in this case (e.g., Adams, 1998, footnote 5, p. 57). However, this leads to wrong results, since then also $p(\sim B|A)=1$, which violates the basic probabilistic principle $p(B|A)+p(\sim B|A)=1$ (see Gilio, 2002, for a discussion). Moreover, from a practical point of view, if $p(B|A)$ is left undefined, counterintuitive consequences may follow. Consider, for example, the following "paradox of the material conditional":

(2) *B*. Therefore, *if A, then B*.

The argument (2) consists of a premise *B* and a conditional as its conclusion. Under the material conditional interpretation, (2) is *logically valid* (i.e., it is *impossible* that the premise set is true while the conclusion is false). However, natural language instantiations can appear counterintuitive (instantiate, for example, "The moon is made of green cheese." for *A* and "The sun will shine in Vienna." for *B*). This mismatch between the logical validity of (2) and counterintuitive instantiations constitute the paradox. From a logical point of view, a logically valid argument remains logically valid whatever the instantiations are. If (2) is formalized in probability logic, however, the paradox is blocked when the conditional is represented by a conditional probability: then, the argument is probabilistically non-informative:

(3) $p(B)=x$. Therefore, $0 \leq p(B|A) \leq 1$ is coherent for all probability values *x*.

An argument is probabilistically non-informative when the premise probabilities do not constrain the probability of the conclusion. More technically, probabilistic non-informativeness means that for all coherent probabilitiy assessments of the premises: the tightest coherent probability bounds on the conclusion coincide with the unit interval, [0,1]. Here, "coherence" means the avoidance of Dutch books. A Dutch book is a combination of symmetric bets which leads to sure win (or to sure loss; see also chapter 4.1. by Hájek and Staffel in this volume). In the coherence-based approach, the avoidance of Dutch books is equivalent to the solvability of a specific linear system. This solvability reflects the existence of at least one probability distribution on a suitable partition of the constituents (i.e., the possible cases), which is compatible with the initial probability assessment. In geometrical terms, a probability assessment can be represented by a prevision point **P** and the set of constituents by a set **Q** of binary points. Then, **P** is coherent if and only if it belongs to the convex hull of **Q**.

If the conditional in (2) is represented by the probability of the material conditional, the paradox is inherited:

(4) *p(B)=x*. Therefore, $x \leq p(\sim A \vee B) \leq 1$ is coherent for all probability values *x*.

Note that in general (3) is probabilistically non-informative for all positive premise probabilities (i.e., $p(B)>0$). For the extreme case $p(B)=1$, when conditional probabilities are defined by (1), then $p(B|A) = 1$ or undefined. This is obvious since if $p(B)=1$, then $p(A\&B)=P(A)$; therefore, by (1), $p(B|A)=p(A\&B)/P(A)=p(A)/P(A)=1$, provided $p(A)>0$. If $p(A)=0$, then $P(B|A)$ is undefined. This result is counterintuitive and does not match the experimental data: people interpret (3) as probabilistically non-informative, even in the case of $p(B)=1$ (Pfeifer and Kleiter, 2011). However, in *coherence-based probability logic* (see, e.g., Coletti and Scozzafava, 2002; Gilio, Pfeifer, and Sanfilippo, 2016), where $p(B|A)$ is primitive and problems with zero antecedent probabilities are avoided, $0 \leq p(B|A) \leq 1$ is coherent even in the extreme case $p(B)=1$ (for a detailed proof see Pfeifer, 2014). This example shows that the evaluation of the rationality of a probabilistic inference depends on whether the underlying probability concept allows for dealing with zero-antecedent probabilities or not. In the framework of coherence, the probabilistic-non-informativeness of argument (3) holds for all probability values of the premises; for approaches, however, which are based on (1), it holds only for positive probabilities.

**From truth-table tasks to probabilistic truth-table tasks**

Logic dominated the psychology of deductive reasoning as a rationality framework in the 20$^{th}$ century. Prominent examples are Braine and O'Brien's "mental logic" (1998; see also chapter 3.2. by O' Brien in this volume) or the "mental rule theory" by Rips (1994) and Johnson-Laird's (1983) theory of mental models (see chapter 2.3. by Johnson-Laird in this volume). The rationality framework of the former two theories is derived from classical logical proof theory ("rule-based"), whereas the latter one is based on logical model theory ("semantic-model-based"). According to these logic-based rationality frameworks people are rational, if they use logically valid rules of inference (like modus ponens) or if they build mental models which are inspired by truth tables. With the advent of the "new paradigm psychology of reasoning", which is characterized by using probabilistic rationality frameworks instead of logic, not only the evaluation of the rationality of human inference changed but also the task paradigms were adapted. The above-mentioned truth table task, for example, became a *probabilistic* truth table task (short PTTT) to investigate how people interpret conditionals (Evans et al., 2003; Oberauer and Wilhelm, 2003; see also chapter 4.6. by Oberauer and Pessach in this volume). From a probability logical point of view, the PTTT presented the following premises to the participants:

(5) $p(A\&B)=x_1$, $p(A\&\sim B)=x_2$, $p(\sim A\&B)=x_3$, and $p(\sim A\&\sim B)=x_4$.

Then the participants were asked to infer their degree of belief in the conditional "*if A, then B*" based on the probabilistic information given in (5). The main result of this task was that most participants responded by values obtained from $x_1/(x_1+x_2)$, which corresponds to the conditional probability interpretation of conditionals ($p(B|A)$; this is consistent with the "Ramsey test" as described in chapter 6.2. by Over and Cruz in this volume). A significant minority responded by $x_1$, which corresponds to the conjunction interpretation of conditionals ($p(A\&B)$). Under the material conditional interpretation, one would expect $x_1+x_3+x_4$ as the modal response in this task. However, experimental evidence for this hypothesis was negligible. When the task was given several times to the same participants, a "shift of interpretation" was observed among those participants who did not use conditional probability responses in the first PTTT tasks. These participants "shifted" to the conditional probability interpretation through the course of the experiment. In the last tasks of the experiment, about 80% of the responses were consistent with the conditional probability interpretation. This is a strong indicator that conditional probability is the competence response (see Fugard et al., 2011; Pfeifer, 2013).

Mostly indicative conditionals ("*if—then*" formulations) with "abstract" material were used in the PTTT (like "if the figure shows a *square*, then the figure is *red*"). Interestingly, the finding that conditional probability is the best predictor for the data was also replicated for a bigger variety of conditionals: causal conditionals ("if *cause*, then *effect*"), counterfactual conditionals ("if *A were the case*, then *B would be the case*"; Over et al., 2007; Pfeifer and Stöckle-Schobel, 2015), and abductive conditionals ("if *effect*, then *cause*"; Pfeifer and Tulkki, 2017a).

The next sections explain why probability logic validates basic connexive principles and why it is nonmonotonic under the conditional probability interpretation of conditionals.

**Probability logic is connexive**

Connexive logics are motivated by the idea that there should be some connection between antecedents and consequents of conditionals in the sense that they should not contradict each other. Connexive logics are alternatives to (classical) logic as they are neither contained in nor proper extensions of it (for an overview see Wansing, 2016). They include, for example, Aristotle's theses:

    (AT1) It is not the case that: if ~$A$, then $A$.

and

(AT2) It is not the case that: if *A*, then ~*A*.

Under the material conditional, (AT1) and (AT2) are contingent in logic (i.e., (AT1) is logically equivalent to ~*A* and (AT2) is logically equivalent to *A*). Thus, within logic, it is rational to say that it depends on the truth value of *A* whether (AT1) and (AT2) is true. Indeed, (AT1) and (AT2) are not tautologies in logic. Experimental data suggest, however, that people believe that (AT1) as well as (AT2) must be true (Pfeifer, 2012). Probability logic allows for validating the rationality of (AT1) and (AT2). First, look at the conditionals (in terms of conditional probabilities): by coherence, $p(A|\sim A)$ and $p(\sim A|A)$ must be equal to zero. Second, these conditionals are negated by negating their consequents, i.e., $p(\sim A|\sim A)$ and $p(\sim\sim A|A)=p(A|A)$, respectively. Since probability one is the only coherent assessment for $p(\sim A|\sim A)$ and for $p(A|A)$, (AT1) and (AT2) are validated. This matches the experimental data (Pfeifer, 2012; Pfeifer and Tulkki, 2017b).

Boethius theses are another instance of connexive principles. Like Aristotle's theses, Boethius theses can be justified within probability logic (but not within logic). The two versions of Boethius theses are (the arrow denotes a conditional):

(BT1) $(A \rightarrow B) \rightarrow \sim(A \rightarrow \sim B)$

and

(BT2) $(A \rightarrow \sim B) \rightarrow \sim(A \rightarrow B)$.

By the narrow scope negation interpretation of negating conditionals,[3] the antecedent of (BT1) is interpreted in probability logic by $P(B|A)$ and its consequent by $P(\sim\sim B|A)$, which is equal to $P(B|A)$. Thus, (BT1) holds in probability logic. Analogously, (BT2) is validated in probability logic. Under the material conditional interpretation, neither (BT1) nor (BT2) hold in general: (BT1) and (BT2) are logically equivalent to *A*. Moreover, Abelard's First Principle, which is another connexive principle, can be rationally justified in probability

---

3   For a wide-scope negation interpretation of negating conditionals see Gilio, Pfeifer, & Sanfilippo (2016) and Pfeifer & Sanfilippo (2017).

logic:

(AFP) ~((A →B) & (A → ~ B))

Since, in general $P(B|A)+P(\sim B|A)=1$, it cannot be the case that both, $P(B|A)$ and $P(\sim B|A)$ are "high" (i.e., at least greater than .5). Therefore, (AFP) is validated in probability logic. However, (AFP) is logically equivalent to *A* under the material conditional interpretation. Aristotle's and Boethius' theses, and Abelard's First Principle are intuitively plausible principles which hold in connexive and in probability logic, but not in (propositional) logic. Aristotle's theses received strong experimental support (Pfeifer, 2012; Pfeifer and Tulkki, 2017b). For the other connexive principles future empirical research is needed.

**Probability logic is nonmonotonic**

Nonmonotonic reasoning is about retracting conclusions in the light of new evidence. For example, from "if this animal is a bird (*B*), then this animal can fly (*F*)" one would not want to conclude that "if this animal is a bird and a penguin (*B&P*), then this animal can fly (*F*)". Logic, however, is monotonic: adding premises to a logically valid argument can only lead to an increase but never to a decrease of the conclusion set (for an overview see, e. g., Antoniou, 2010; Strasser and Antonelli, 2016). Therefore, conclusions cannot be retracted in logic. Under the material conditional interpretation, the above-mentioned argument is logically valid. The argument form is called "monotonicity" (or "premise strengthening"):

(MON) ~*B* v *F* logically implies ~(*B&P*) v *F*.

In probability logic, however, the corresponding argument form is probabilistically non-informative and monotonicity is therefore blocked:

(6) $p(F|B)=x$. Therefore, $0 \leq p(F|B\&P) \leq 1$ is coherent for all probability values *x*.

Basic rationality principles for retracting conclusions in the light of new evidence are concentrated around System P (Kraus et al., 1990). The principles of System P are considered as minimal rationality requirements for any system of nonmonotonic reasoning: it is therefore a key system for reasoning in general. Various different semantics were developed for nonmonotonic reasoning systems, among which some are probability logical ones (see, e. g., Adams, 1975; Goldszmidt and Pearl, 1996; Schurz, 1996; Gilio, 2002; Hawthorne and Makinson, 2007). Psychologically, Pfeifer and Kleiter (2005, 2006, 2010) and Pfeifer and Tulkki (2017b) present experimental data supporting the coherence-based probability semantics of System P (Gilio, 2002; for experimental studies on the possibilistic semantics of System P see, e. g., Da Silva Neves et al., 2002; Benferhat et al., 2005). In the coherence semantics default conditionals "*A normally B*" are interpreted as coherent conditional probabilities, which may be imprecise (i. e., interval-valued probabilities). For each rule of System P, Gilio (2002) proved the probability propagation rules, which describe how the probabilities of the premises are propagated to the conclusion. As an example, consider the *and rule*:

(AND) $x' \leq p(B|A) \leq x''$ and $y' \leq p(C|A) \leq y''$. Therefore,

$\max\{0, x'+y'-1\} \leq p(B\&C|A) \leq \min\{x'', y''\}$ is coherent.

It can easily be seen that even in cases where the premise probabilities are point-valued (i.e., $x' = x''$ and $y' = y''$), the probability of the conclusion is usually interval-valued. In the extreme case, where the premise probabilities are equal to one, the only coherent conclusion probability is also equal to one. Experimental data suggest that most people infer coherent interval responses. The majority of those people who violate coherence violate the lower bound (Pfeifer & Kleiter, 2005). These data speak also against the common misunderstanding that people are unable to perform probabilistic reasoning because of the high frequency of "conjunction fallacies" allegedly committed in Tversky and Kahneman's (1983) well-known Linda task. The conjunction fallacy consists in ranking the probability of a conjunction

(*B&C*) as higher compared to one of its conjuncts (*C*). In the context of (AND), this would mean that the *upper* probability bound on the conclusion is violated. In the experimental data on (AND), however, those people who violated the coherent interval, violated the *lower* probability bound (Pfeifer & Kleiter, 2005). Even if the terms "and" and "probability" are mentioned, it does not mean that actual conjunction probabilities are investigated in the Linda task: the participants might not interpret the task as a task about conjunction probabilities. Concerning the other rules of System P, strong agreement between the participants' interval responses and the coherent intervals were observed (Pfeifer and Kleiter, 2005, 2006, 2010; Pfeifer and Tulkki, 2017b). Moreover, the majority of the participants correctly understood that (6) and that contraposition (e.g., $p(B|A) = x$. Therefore, $0 \leq p(\sim A|\sim B) \leq 1$ is coherent for all probabilities $x$) are probabilistically non-informative. This is interesting as these argument forms are logically valid under the material conditional but they cannot be validated without further assumptions in a nonmonotonic reasoning system. Adding monotonicity or contraposition to System P, for example, would make System P monotonic (which is undesirable of course). Transitivity is also probabilistically non-informative (i.e., $p(B|A) = x$, $p(C|B)=y$. Therefore, $0 \leq p(C|A) \leq 1$ is coherent for all probabilities $x$ and $y$) and its addition to System P would make it monotonic. Experimental data suggest, that people interpret the task material of Transitivity presumably because of pragmatic reasons as CUT of System P (Pfeifer & Kleiter, 2006, 2010): CUT or cumulative transitivity strengthens the premises of transitivity by adding the antecedent of the first premise by conjunction to the antecedent of the second premise (Gilio, 2002):

(CUT) $p(B|A) = x$, $p(C|A\&B)=y$. Therefore, $xy \leq p(C|A) \leq xy+1-x$.

Note that the probability propagation rules of (CUT) coincide with those of the probabilistic modus ponens (Pfeifer & Kleiter, 2006a):

(MP) $p(B) = x$, $p(C|B)=y$. Therefore, $xy \leq p(C) \leq xy+1-x$.

This close relationship between (CUT) and (MP) is explained by the fact that unconditional probabilities are defined in probability logic by the following principle:

(7) $p(A)=$def. $p(A|verum)$, where *"verum"* denotes a logical tautology.

By replacing *A* by *verum* in (CUT) and by (7), we obtain (MP). Modus ponens is one of the most frequently investigated argument forms in the psychology of reasoning. Its non-probabilistic version is usually endorsed by most participants (Evans et al., 1993). The clear majority of responses in tasks on the probabilistic modus ponens support the predictions by probability logic (Pfeifer and Kleiter, 2009; Pfeifer and Tulkki, 2017b).

**Concluding remarks**

This chapter characterized probability logic as a generalization of logic and explained the importance of zero-antecedent probabilities. Probability logic is connexive and nonmonotonic. It is a powerful tool to investigate the rationality of reasoning in a unified and systematic way. Experimental studies support its descriptive validity. Future normative and descriptive research directions should include nested and compound conditionals. The probabilistic modus ponens and other argument forms, for example, were recently generalized to deal with nested and compounds of conditionals (Sanfilippo et al., 2017; Sanfilippo et al., 2018). Interestingly, the uncertainty propagation rules for the modus ponens involving *nested conditionals* coincide with those of the non-nested modus ponens (MP). For instance, consider the following nested version of the modus pones: from *the cup breaks if dropped* ($D{\rightarrow}B$) and *if the cup breaks if dropped, then the cup is fragile* (($D{\rightarrow}B){\rightarrow}F$) infer *the cup is fragile* (*F*). Here, the lower bound of the degree of belief in the conclusion *F* equals the product of the degrees of belief in the premises and the upper bound on *F* equals the sum of the lower bound on *F* plus 1 minus the degree of belief in $D{\rightarrow}B$, which coincides with the uncertainty propagation rules of the non-nested (MP) (for details see Sanfilippo et al., 2017).

In this approach, which is based on previsions in conditional random quantities, the law of import-export does not hold (Gilio and Sanfilippo, 2014), which is key to block Lewis' (1976) notorious triviality results. Lewis' triviality results show that sentences like $(D{\rightarrow}B){\rightarrow}F$ must not be simply interpreted by $p(F|(B|D))$. Rather, a richer formal structure is required for properly investigating such structures. Future work is needed to assess the psychological plausibility of this approach.

Finally, one might wonder why various non-classical logics—which validate inutitively plausible rationality principles—were broadly neglected in the psychological literature, even if they were available already for decades. The reasons might be due to research traditions. This chapter proposed probability logic as normatively and descriptively appealing rationality framework for human reasoning, which combines (i) the requirement of plausible *qualitative* logical principles (like nonmonotonicity and connexivity) with (ii) the expressibility of *quantitative* degrees of beliefs for investigating reasoning and argumentation under uncertainty.

References

Adams, E.W. (1975). *The logic of conditionals*. Dordrecht: Reidel.

Adams, E.W. (1998). *A primer of probability logic*. Stanford: CSLI.

Antoniou, G. (1997). *Nonmonotonic reasoning*. Cambridge: MIT Press.

Braine, M.D.S., & O'Brien, D.P. (Eds.) (1998). *Mental logic*. Mahwah: Erlbaum.

Coletti, G. & Scozzafava, R. (2002). *Probabilistic logic in a coherent setting*. Dordrecht: Kluwer.

Coletti, G., Scozzafava, R., & Vantaggi, B. (2001). Probabilistic reasoning as a general unifying tool. In Benferhat, S. & Besnard, P. (Eds.). *Symbolic and Quantitative Approaches to Reasoning with Uncertainty* (pp. 120–131). Cham: Springer (LNAI, vol. 2143).

Demey, L., Kooi, B. & Sack, J. (2017). Logic and probability. In E. N. Zalta (Ed.), *The Stanford Encyclopedia of Philosophy* (Summer 2017 ed.).

Elqayam, S., Bonnefon, J.-F., & Over, D.E. (Eds.) (2016). *New paradigm psychology of reasoning*. London: Routledge.

Evans, J.St.B.T., Newstead, S.E., & Byrne, R.M.J. (1993). *Human reasoning*. Hove: Lawrence Erlbaum.

Evans, J.St.B.T., Handley, S.J., & Over, D.E. (2003). Conditionals and conditional probability. *Journal of Experimental Psychology: LMC 29*(2), 321–355.

Fugard, A.J.B., Pfeifer, N., Mayerhofer, B., & Kleiter, G.D. (2011). How people interpret conditionals: shifts towards the conditional event. *Journal of Experimental Psychology: LMC 37*(3), 635–648.

Gilio, A. (2002). Probabilistic reasoning under coherence in System P. Annals of Mathematics and Artificial Intelligence 34, 5–34.

Gilio, A., Pfeifer, N., & Sanfilippo, G. (2016). Transitivity in coherence-based probability logic. *Journal of Applied Logic 14*, 46–64.

Gilio, A. & Sanfilippo, G. (2014). Conditional random quantities and compounds of conditionals. *Studia Logica 102*(4), 709–729.

Goldszmidt, M. & J. Pearl (1996). Qualitative probabilities for default reasoning, belief revision, and causal modeling. *Artificial Intelligence 84*, 57–112.

Haenni, R., Romeijn, J.-W., Wheeler, G., & Williamson, J. (2011). *Probabilistic logics and probabilistic networks*. Dordrecht: Springer.

Hailperin, T. (1996). *Sentential probability logic*. Bethlehem: Lehigh University Press.

Hailperin, T. (2011). *Logic with a probability semantics*. Bethlehem: Lehigh University Press.

Hawthorne, J. & Makinson, D. (2007). The quantitative/qualitative watershed for rules of uncertain inference. *Studia Logica 86*, 247–297.

Johnson-Laird, P. N. (1983). *Mental models*. Cambridge: Cambridge University Press.

Kleiter, G.D., Fugard, A.J.B., & Pfeifer, N. (2018). A process model of the understanding of uncertain conditionals. *Thinking & Reasoning 24*(3), 386–422.

Kraus, S., Lehmann, D., & Magidor, M. (1990). Nonmonotonic reasoning, preferential models and cumulative logics. *Artificial Intelligence 44*, 167–207.

Lewis, D. (1976). Probabilities of conditionals and conditional probabilities. *Philosophical Review 85*, 297–315.

Oaksford, M. & Chater, N. (Eds.) (2010). *Cognition and conditionals.* Oxford: Oxford University Press.

Oberauer, K. & Wilhelm, O. (2003). The meaning(s) of conditionals: Conditional probabilities, mental models and personal utilities. *Journal of Experimental Psychology: LMC 29*(4), 680–693.

Over, D.E. & Baratgin, J. (2017). The "defective" truth table: Its past, present, and future. In Galbraith, N., Lucas, E., & Over, D.E. (Eds.), *The thinking mind: A Festschrift for Ken Manktelow* (pp. 15–28). Hove: Psychology Press.

Over, D.E., Hadjichristidis, C., Evans, J.St.B.T., Handley, S.J., & Sloman, S. (2007). The probability of causal conditionals. *Cognitive Psychology 54*, 62– 97.

Pfeifer, N. (2012). Experiments on Aristotle's Thesis: towards an experimental philosophy of conditionals. *The Monist 95*(2), 223–240.

Pfeifer, N. (2013). The new psychology of reasoning: A mental probability logical perspective. *Thinking & Reasoning 19*(3–4), 329–345.

Pfeifer, N. (2014). Reasoning about uncertain conditionals. *Studia Logica 102*(4), 849–866.

Pfeifer, N. & Kleiter, G.D. (2005). Coherence and nonmonotonicity in human reasoning. *Synthese 146*(1-2), 93–109.

Pfeifer, N. & Kleiter, G.D. (2006a). Inference in conditional probability logic. *Kybernetika 42*, 391–404.

Pfeifer, N. & Kleiter, G.D. (2006b). Is human reasoning about nonmonotonic conditionals probabilistically coherent? In *Proceedings of the 7th Workshop on Uncertainty Processing, Mikulov* (pp. 138–150).

Pfeifer, N. & Kleiter, G.D. (2009). Framing human inference by coherence based probability logic. *Journal of Applied Logic 7*(2), 206–217.

Pfeifer, N. & Kleiter, G.D. (2010). The conditional in mental probability logic. In Oaksford, M. and Chater, N. (Eds.), *Cognition and conditionals* (pp. 153–173). Oxford: Oxford University Press.

Pfeifer, N. & Kleiter, G.D. (2011). Uncertain deductive reasoning. In Manktelow, K. Over, D.E., & Elqayam, S. (Eds.), *The science of reason: a Festschrift for Jonathan St. B.T. Evans* (pp. 145–166). Hove: Psychology Press.

Pfeifer, N. & Sanfilippo, G. (2017). Probabilistic squares and hexagons of opposition under coherence. *International Journal of Approximate Reasoning 88*, 282–294.

Pfeifer, N. & Sanfilippo, G. (2019). Probability propagation in selected Aristotelian syllogisms. In Kern-Isberner, G. & Ognjanović, Z. (Eds.), *Symbolic and Quantitative Approaches to Reasoning with Uncertainty* (pp. 419–431). Cham: Springer (LNCS, vol. 11726).

Pfeifer, N. & Stöckle-Schobel, R. (2015). Uncertain conditionals and counterfactuals in (non-)causal settings. In G. Arienti et al. (Eds.), *Proceedings of the EuroAsianPacific Joint Conference on Cognitive Science* (pp. 651–656). Volume 1419. Aachen: CEUR Workshop Proceedings.

Pfeifer, N. & Tulkki, L. (2017a). Abductive, causal, and counterfactual conditionals under incomplete probabilistic knowledge. In Gunzelmann, G. et al. (Eds.), *Proceedings of the 39th CogSci Meeting* (pp. 2888–2893). Austin: The Cognitive Science Society.

Pfeifer, N. & Tulkki, L. (2017b). Conditionals, counterfactuals, and rational reasoning. An experimental study on basic principles. *Minds and Machines 27*(1), 119–165.

Rips, L. J. (1994). *The psychology of proof: deductive reasoning in human thinking*. Cambridge: MIT Press.

Sanfilippo, G., Pfeifer, N., & Gilio, A. (2017). Generalized probabilistic modus ponens. In Antonucci, A., Cholvy, L., & Papini, O. (Eds.), *Symbolic and Quantitative Approaches to Reasoning with Uncertainty* (pp. 480–490). Cham: Springer (Volume 10369 of LNCS).

Sanfilippo, G., Pfeifer, N., Over, D.E., & Gilio, A. (2018). Probabilistic inferences from conjoined to iterated conditionals. *International Journal of Approximate Reasoning 93*, 103–118.

Schurz, G. (1997). Probabilistic default reasoning based on relevance and irrelevance assumptions. In Gabbay, D. et al. (Ed.), *Qualitative and Quantitative Practical Reasoning*

(pp. 536–553). Berlin: Springer (Volume 1244 of LNAI).

Strasser, C. & Antonelli, G.A. (2016). Non-monotonic logic. In Zalta, E.N. (Ed.), *The Stanford Encyclopedia of Philosophy* (Winter 2016 ed.).

Tversky, A. & Kahneman, D. (1983). Extensional versus intuitive reasoning: the conjunction fallacy in probability judgment. *Psychological Review 90*, 293–315.

Wansing, H. (2016). Connexive logic. In Zalta, E.N. (Ed.), *The Stanford Encyclopedia of Philosophy* (Spring 2016 ed.).

Wason, P.C. & Johnson-Laird, P.N. (1972). *The psychology of reasoning*. Cambridge: Harvard University Press.